\documentclass{article}

\usepackage[preprint]{corl_2026} 

\usepackage{graphicx} 
\usepackage{amsmath} 
\usepackage{amssymb}  
\usepackage{caption}
\usepackage{hyperref}
\usepackage{algorithm}
\usepackage{algorithmic}
\usepackage{multirow}
\usepackage{booktabs}
\usepackage{pifont}
\usepackage{enumitem}
\usepackage{makecell}
\usepackage{multicol}
\usepackage{wrapfig}
\usepackage[title]{appendix}
\usepackage{graphicx}
\usepackage{subcaption}

\makeatletter
\renewcommand{\paragraph}{%
  \@startsection{paragraph}{4}{\z@}%
                {0.1ex \@plus 0.1ex \@minus 0.1ex}%
                {-1em}%
                {\normalsize\bf}%
}
\makeatother

\newcommand{\cmark}{\ding{51}}%
\newcommand{\xmark}{\ding{55}}%

\title{EgoGuide: Egocentric Guidance for Efficient Robot-Free Demonstration Collection and Learning}

\author{
Yue Xu\textsuperscript{1}, Mingtao Nie\textsuperscript{1}, Tianle Li\textsuperscript{1}, Hong Li\textsuperscript{1}, Yibo Luo\textsuperscript{1}, Siyuan Huang\textsuperscript{3}, Yong-Lu Li\textsuperscript{1,2}\thanks{Corresponding author.}
\\
\textsuperscript{1}Shanghai Jiao Tong University,
\textsuperscript{2}Shanghai Innovation Institute\\
\textsuperscript{3}Beijing Institute for General Artificial Intelligence (BIGAI)\\
\texttt{\selectfont\{silicxuyue, yonglu\_li\}@sjtu.edu.cn}
}

\begin{document}
\maketitle

\begin{abstract}
    Robot learning from real-world demonstrations is currently constrained by data scaling. Universal Manipulation Interface (UMI) provides an efficient robot-free data collection interface, yet current UMI-style pipelines often collect redundant demonstrations and lack global scene context. To improve data efficiency, we present EgoGuide, a collection interface that records synchronized wrist and head/egocentric observations and couples them with online visual-geometric data quality guidance.
    We also introduce a Gated Egocentric Residual Policy for robust learning from a viewpoint-varying egocentric camera, allowing head/egocentric context to correct ambiguous local observations while preserving stable wrist-view control. Real-world experiments show that EgoGuide reduces the required number of data episodes and improves data efficiency. The residual policy further improves robustness under visual occlusion.\\
    \textbf{Project Page:} \url{https://silicx.github.io/EgoGuide/} 
\end{abstract}


\begin{figure}[h]
    \includegraphics[width=\linewidth]{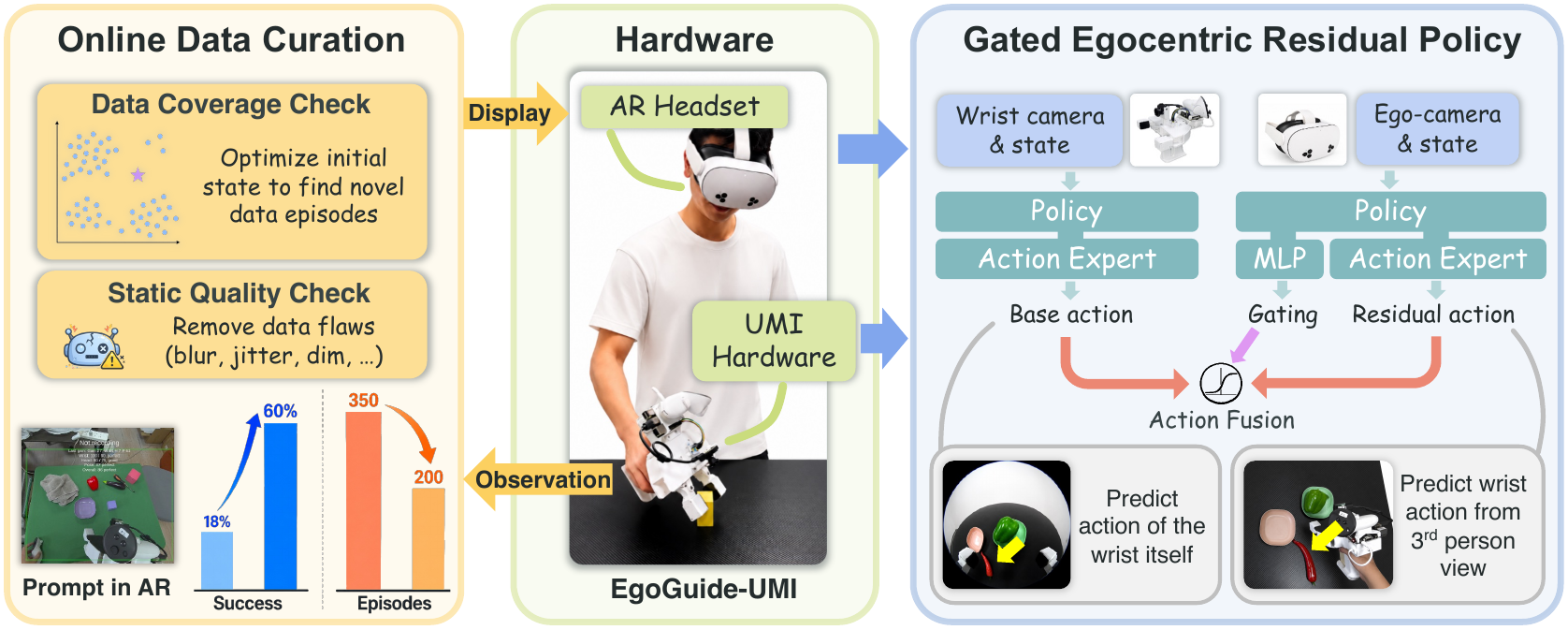}
    \caption{
    Overview of EgoGuide. We extend UMI-style data collection with synchronized wrist and head/egocentric sensing and online visual-geometric data quality guidance through AR feedback. A gated residual policy uses the head/egocentric view to complement a stable wrist-view policy.
    }
    \label{fig:teaser}
    \vspace{-3mm}
\end{figure}

\section{Introduction}

Currently, the scaling of robot learning remains fundamentally constrained by real-world demonstration data.
Universal Manipulation Interface (UMI)~\citep{umi} offers an appealing alternative to costly teleoperation systems and shows a promising direction for scalable manipulation data collection.
However, scaling robot learning with robot-free interfaces such as UMI is not simply a matter of recording more trajectories.
We define \textit{data efficiency} as the human collection effort required to train a policy to a target success rate.
From this perspective, UMI can be less data-efficient than teleoperation systems, sometimes requiring more than $5\times$ episodes for comparable success.

We identify two critical bottlenecks.
First, UMI data are collected with a gripper surrogate and transferred to a robot with different kinematics, coordinates, and execution constraints, so policies require broad coverage of data variation rather than redundant successful trajectories; yet demonstrators receive little feedback about which states are already covered or underrepresented.
Second, because most UMI systems rely on a single wrist camera, observations can be too local to capture the full task state under occlusion, object disappearance, or long-horizon progress.
These bottlenecks suggest that improving UMI data efficiency requires both collection-time coverage guidance and the exploitation of complementary egocentric observations.

To address these issues, we first present \emph{EgoGuide}, an integrated robot-free data collection system that combines multi-view demonstration collection and online data quality guidance.
EgoGuide targets the data diversity and coverage bottleneck by making the demonstrator aware of the data quality of the current visual-geometric state.
We augment the handheld UMI interface with AR-based egocentric sensing and spatial tracking. Multiple modalities stream synchronously to a workstation, which computes an online data coverage score from wrist and head/egocentric visual-geometric information.
The score is rendered in the AR interface to encourage adjustments to the initial hand pose, object arrangement, viewpoint, or workspace configuration before recording an episode.
EgoGuide also allows and encourages recording from the middle of a task, extending data quality control beyond the early stages.

For robust egocentric policy learning, we introduce the \emph{Gated Egocentric Residual Policy} (GERP), which addresses the observability bottleneck by using head/egocentric context to correct wrist-view control \textit{on demand}.
We argue that human head movement is often unintentional, so rather than building an active egocentric camera with imitation learning~\citep{xiong2025vision,zou2026activeglasses,zeng2025activeumi}, we propose to learn a robust policy that can generalize to a fixed egocentric camera, aligning with realistic robot deployment.
GERP keeps a wrist-view policy as a stable base, conditions an egocentric branch on the head/egocentric image and the wrist pose relative to the head frame, and gates the egocentric action candidate with the base action.
This preserves wrist-view control in ordinary states while allowing head/egocentric context to modify the action when local observations are ambiguous or incomplete.

We evaluate on real-world manipulation tasks covering standard UMI-style settings as well as failure cases involving occlusion and limited visual coverage.
On regular long-horizon tasks, our online collection guidance substantially improves success with the same dataset size and, on Pepper Sorting, reaches comparable success using only \textbf{50\%} as many demonstrations.
The egocentric residual policy further improves performance under occlusion.
These results support the central hypothesis of this paper: reliable scaling of UMI requires both collecting more informative demonstrations and learning to use heterogeneous observations.


\section{Related Work}

\paragraph{Robot-Free Demonstration Collection.}
Large-scale robot data collection remains a bottleneck for robot learning. Teleoperation systems such as ALOHA~\citep{zhao2023learning}, Mobile ALOHA~\citep{fu2024mobile}, and GELLO~\citep{wu2024gello} improve the accessibility of robot-native data collection.
Recently, Universal Manipulation Interface (UMI)~\citep{umi} introduced a portable gripper-like device that allows low-cost in-the-wild data collection. Subsequent works extend this paradigm. FastUMI improves portability and collection throughput~\citep{zhaxizhuoma2025fastumi}. exUMI~\citep{exumi}, TacUMI~\citep{cheng2026tacumi}, ViTaMIn~\citep{liu2025vitamin}, and FreeTacMan~\citep{wu2025freetacman} incorporate tactile or force-related sensing for contact-rich manipulation. MV-UMI~\citep{rayyan2025mv} and UMI-3D~\citep{wang2026umi} enhance visual or spatial observability. DexUMI~\citep{xu2025dexumi}, HoMMI~\citep{xu2026hommi}, and HuMI~\citep{nai2026humanoid} extend toward dexterous or whole-body manipulation.
Building on this, we study how to improve demonstration quality through online guidance.

Egocentric vision is also exploited in UMI systems. Vision in Action (ViA)~\citep{xiong2025vision} first proposes to record UMI data with an additional egocentric camera and learns an active robot head policy from human demonstrations.
ActiveUMI~\citep{zeng2025activeumi} also introduces head-mounted observations with active head deployment. EgoMI~\citep{yu2025egomi} studies active egocentric perception in whole-body manipulation. Unlike methods that directly model or execute active head motion, we use egocentric observations as auxiliary context in a fixed egocentric camera configuration.

\paragraph{Robot Data Quality Evaluation.}
Imitation learning is sensitive not only to dataset scale, but also quality. Interactive imitation-learning methods such as DAgger~\citep{ross2011reduction}, SafeDAgger~\citep{zhang2016query}, and HG-DAgger~\citep{kelly2019hg} collect additional supervision in states visited by the learned policy. 
RoboPocket~\citep{fang2026robopocket} brings the DAgger strategy to the UMI system for online data curation.
Offline studies such as RoboMimic~\citep{mandlekar2021matters}, RoboTurk~\citep{mandlekar2018roboturk}, and RoboNet~\citep{dasari2019robonet} further show that policy performance depends on demonstration quality and diversity.
DemInf~\citep{hejna2025robot} estimates trajectory utility from state diversity and action predictability. Demo-SCORE~\citep{chen2025curating} uses online robot experience to identify useful demonstrations. CUPID~\citep{agia2025cupid} estimates the contribution of individual demonstrations using influence functions. These methods filter or select demonstrations after data collection, while we enable guidance \textbf{during} robot-free data collection.


\section{EgoGuide System}

\begin{figure}[t]
    \centering
    \includegraphics[width=\linewidth]{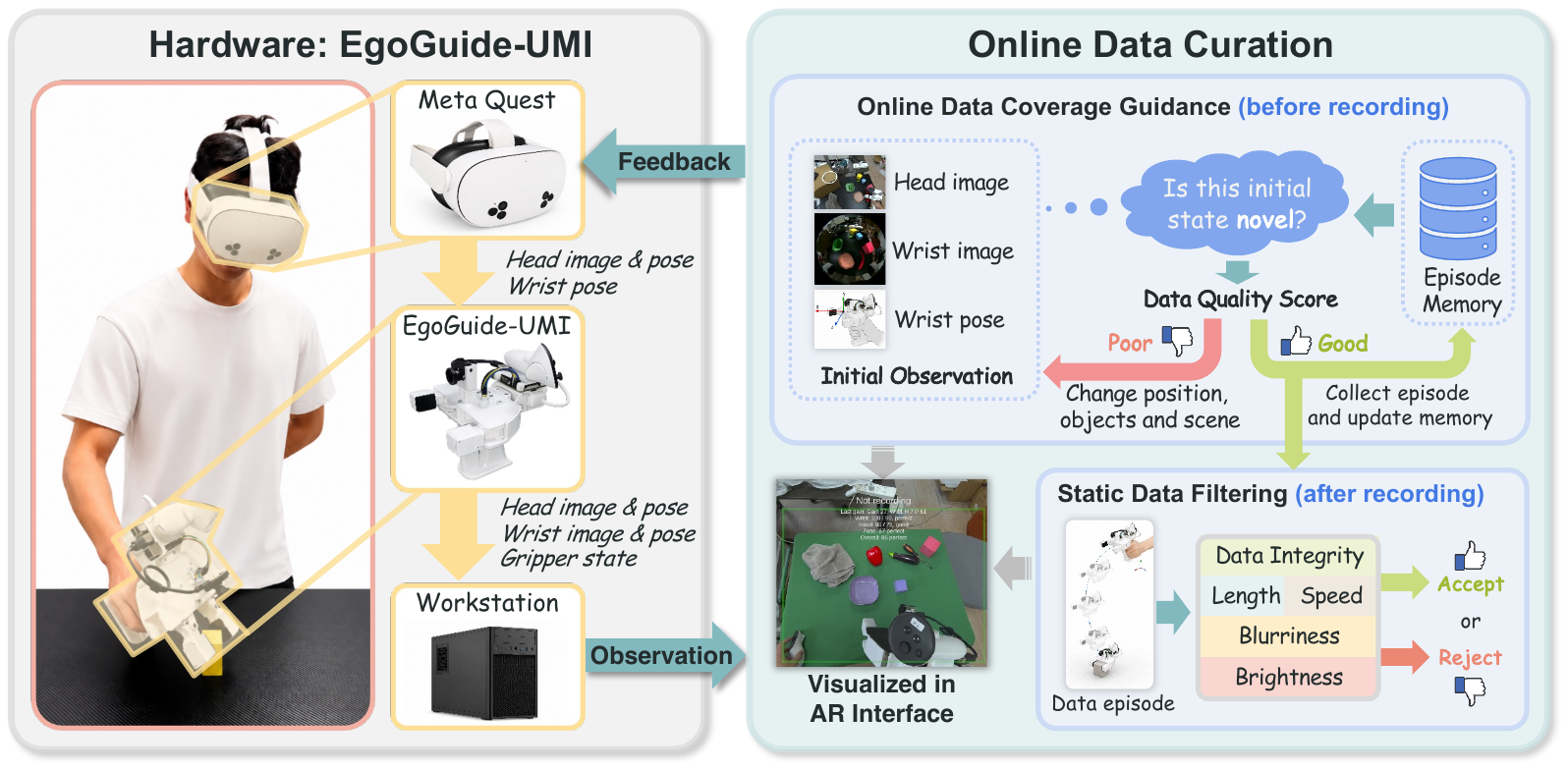}
    \caption{Overview of the EgoGuide hardware system. EgoGuide-UMI records wrist and head/egocentric views, gripper proprioception, and spatial poses, while the online curation module computes coverage scores and sends AR guidance back to the demonstrator.}
    \label{fig:hardware}
\end{figure}

EgoGuide is designed as a closed-loop robot-free collection system, including both UMI hardware design and an online data curation method. Fig.~\ref{fig:hardware} summarizes the hardware and guidance pipeline.

\subsection{EgoGuide-UMI Hardware}
\label{sec:umi-hardware}

We first propose our hardware design \textbf{EgoGuide-UMI}, based on the open-source exUMI ~\citep{exumi}.

\paragraph{Handheld gripper interface.}
EgoGuide-UMI follows the UMI-style principle of collecting robot-free demonstrations with an easy-to-fabricate hand-held gripper. A rotary sensor mounted on the gripper joint measures the opening width $g$ and sends it to an on-device Raspberry Pi controller. We mount an industrial board-level fisheye camera near the gripper to obtain the wrist image $I^W$ so that the wrist stream is captured directly by the controller with lower latency.

\paragraph{Egocentric sensing and spatial tracking.}
We use a Meta Quest headset for head/egocentric sensing and spatial tracking. It provides the head-camera image $I^H$ through the Unity Passthrough API and the head pose $T^H$; throughout the paper, \(H\) denotes this head-mounted egocentric camera. A Quest controller attached to the gripper provides the wrist pose $T^W$. The AR passthrough view matches the demonstrator's view, giving global context beyond the local wrist camera, and the controller buttons are used to control recording.

\paragraph{Synchronized wireless streaming.}
We synchronize all modalities over UDP. The headset streams \(I^H\), \(T^H\), \(T^W\), and the recording state at 72~Hz to the Raspberry Pi controller. The controller captures \(I^W\) and gripper width \(g\) at 20~Hz, aligns them to the closest headset timestamp, and sends the combined packet to a workstation. In our setup, the workstation uses an Intel Core i5-13600KF CPU and an NVIDIA RTX 4070 GPU to store episodes, compute online coverage scores, and return AR guidance. The system runs on a standard 1000~M WLAN router with cross-modality synchronization within 20~ms (less than camera exposure time) and end-to-end workstation latency within 100~ms.

Each aligned observation is $o = \{I^W, I^H, T^W, T^H, g\}, \qquad T^W,T^H \in SE(3)$,
where \(I^W\) and \(I^H\) are the wrist-view and head/egocentric images, \(T^W\) and \(T^H\) are the wrist and head poses in the shared collection world frame, and \(g\) denotes the gripper width.

\subsection{Online Data Curation}
\label{sec:data-eval}

Beyond recording demonstrations, EgoGuide uses the live EgoGuide-UMI stream to improve data quality by guiding initial states online and filtering episodes afterward.

\paragraph{Online data coverage guidance.}
Before each episode, EgoGuide estimates whether the current state expands the dataset coverage. Since robust UMI policies rely on diverse demonstrations, we define a \textbf{data coverage score} to quantify how much the current collection state occupies an under-explored region of the existing dataset. This score is computed over three complementary signals. Wrist images reflect the diversity of policy inputs, including hand-object appearance and contact configuration. Wrist poses capture geometric and action-side diversity, since UMI demonstrations supervise end-effector motion and a narrow pose distribution usually induces a narrow action distribution. Head/egocentric images capture object arrangement, workspace layout, and contextual cues from a global viewpoint that may be invisible to the wrist camera.

For image modalities, EgoGuide extracts normalized visual features from the wrist image \(I^W\) and the head/egocentric image \(I^H\), then compares each feature to a view-specific feature memory. For view \(m\in\{W,H\}\) and encoder \(e\), the visual similarity is the average cosine similarity to the \(k\) nearest features in the corresponding memory:
\begin{equation}
    z^{m,e}=\bar{\phi}_{e}(I^m),
    \qquad
    s_{m,e}=
    \frac{1}{k}\sum_{j\in\mathrm{NN}_k(z^{m,e},\mathcal{M}_{m,e})}
    (z^{m,e})^\top z_j^{m,e}.
\end{equation}
We compute this score with both DINOv2~\citep{oquab2023dinov2} and CLIP~\citep{clip} so the guidance signal captures complementary visual cues from local appearance and geometry to object- and scene-level semantics.

For geometric coverage, EgoGuide uses the wrist pose \(T^W\). We represent it by translation and quaternion orientation, and compute a nearest-neighbor pose similarity in a memory \(\mathcal{M}_p\) using a normalized translation-rotation distance. High similarity indicates a repeated wrist configuration, while low similarity indicates an under-explored region of the pose space.

All similarity scores are converted into novelty percentiles with respect to the current memories, so higher values consistently mean higher coverage gain. This normalization is important for user feedback: raw similarity scores have encoder- and modality-dependent ranges, often with limited perceptual contrast, whereas demonstrators mainly need the relative novelty of the current state. We aggregate the two encoder-specific image scores within each view and display three 0--100 guidance values in the AR interface: wrist-view novelty, egocentric-context novelty, and wrist-pose novelty. The estimator runs at 2~Hz, allowing the demonstrator to adjust the initial hand pose, object arrangement, viewpoint, or workspace layout before recording an episode. Since these initial conditions affect approach direction, contact sequence, occlusion patterns, and recovery behavior, the feedback improves trajectory diversity without on-policy rollouts like DAgger~\citep{ross2011reduction}.

\paragraph{Partial demonstration.}
Because initial-state guidance mainly affects early-stage variation, EgoGuide also supports starting a new recording from the middle of a task. The demonstrator can start collection from an underrepresented intermediate state or subtask. This gives demonstrators more flexibility, turns later-stage states into controllable starts, and improves coverage beyond the beginning of the task. We show the efficacy of partial demonstration in Sec.~\ref{sec:result}.

\paragraph{Static data filtering.}
After an episode ends, EgoGuide applies deterministic quality checks before the sample enters the memory. The system rejects an episode if: (1) missing any required modality, (2) is too short, (3) wrist or head motion contains implausible jumps, (4) the images are severely blurred or outside a normal brightness range. Blur is measured by the Laplacian variance of the gray-scale image, and pose discontinuities are detected using linear and angular velocity thresholds. These checks remove sensor failures and physically inconsistent trajectories while avoiding a learned filter that could bias the dataset. Empirically, this stage discards around 2\%--5\% of collected episodes. Accepted episodes are downsampled and added to the memory.

\section{Algorithm}

\begin{figure}[t]
    \centering
    \includegraphics[width=\linewidth]{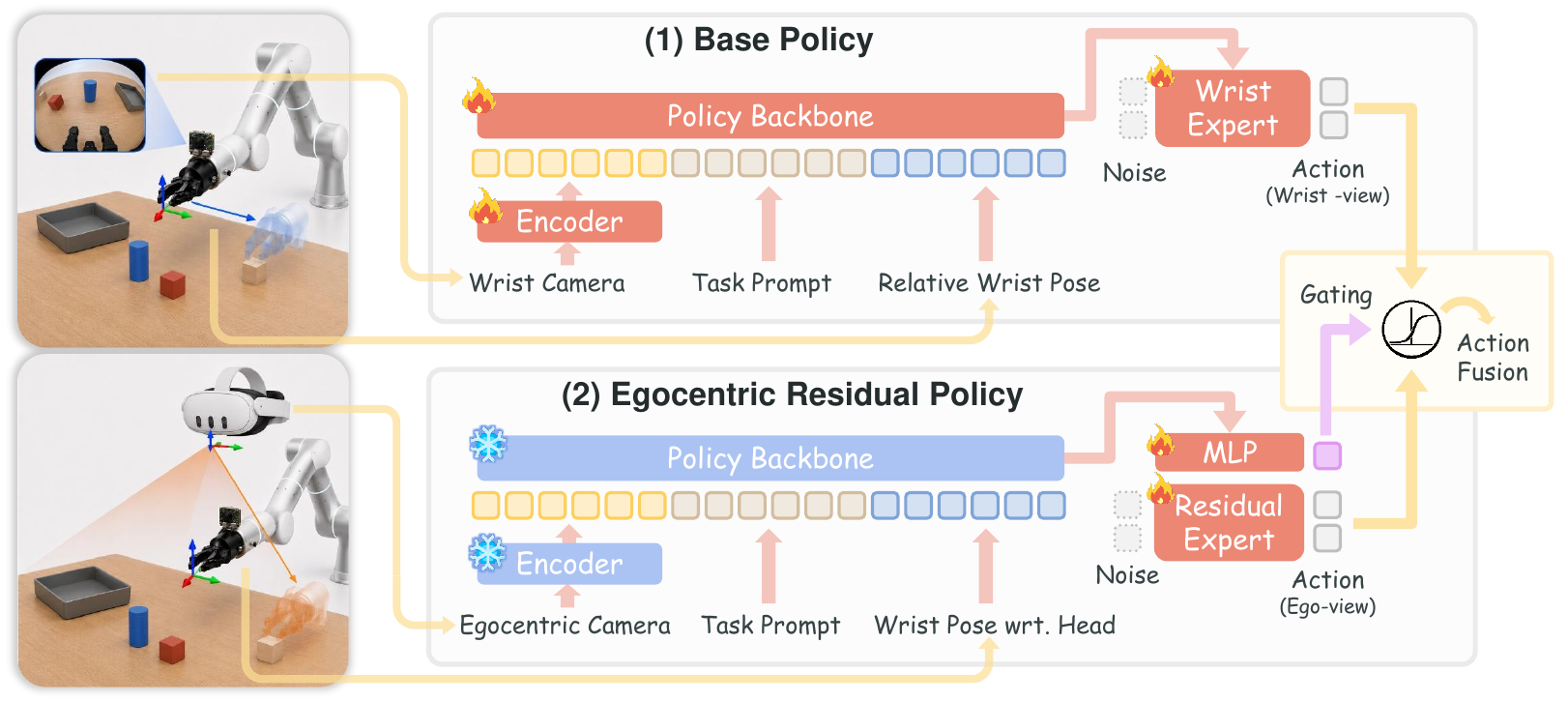}
    \caption{
    Overview of \textbf{G}ated \textbf{E}gocentric \textbf{R}esidual
    \textbf{P}olicy (\textbf{GERP}). A wrist-only base policy predicts a
    nominal action from \(I^W\). An independent egocentric residual branch receives the head/egocentric image \(I^H\) and the wrist pose with respect to the head frame, then predicts an residual action in the same wrist-relative action space. A learned gate blends this candidate with the base action.
    }
    \label{fig:algo}
\end{figure}

Online coverage guidance improves UMI demonstration diversity, but challenging tasks still require a policy that can use the head/egocentric observations collected by EgoGuide. Directly conditioning the full action policy on a moving human viewpoint can be brittle because the head pose is not a controlled robot action, the view can be noisy or redundant, and the visible hand-held device introduces embodiment mismatch.
We therefore propose \textbf{G}ated \textbf{E}gocentric \textbf{R}esidual \textbf{P}olicy (\textbf{GERP}), which treats egocentric perception as a gated alternative action cue rather than a replacement for wrist-view control. The wrist branch provides the default local manipulation behavior, while the egocentric branch proposes a complete action candidate when broader scene context is useful. Unlike active-perception methods that imitate or execute head motion~\citep{xiong2025vision,zeng2025activeumi,yu2025egomi}, GERP does not model the head camera as an action output, and can be deployed with a fixed calibrated egocentric camera. Fig.~\ref{fig:algo} summarizes the design.

\paragraph{Stage 1: wrist-only base policy.}
Let \(\mathbf{A}^\star\) denote the ground-truth demonstration action chunk used as supervision. Following the UMI convention, it is expressed in a wrist-relative action space to the current wrist pose, together with the gripper command. All predicted actions including the base action \(\mathbf{A}^b\), the egocentric action candidate \(\mathbf{A}^r\), and the fused action \(\hat{\mathbf{A}}\), use this same action space.

In the first stage, we train a standard wrist-view diffusion policy. Given the wrist image, the current wrist pose, and the task instruction \(\ell\), the base policy predicts a nominal action chunk:
\begin{equation}
    \mathbf{A}^b
    =
    \pi_b(I^W,T^W,\ell),
    \label{eq:base_policy}
\end{equation}
where the recorded wrist pose \(T^W\) is expressed in the shared collection/world frame. The base policy is trained with the standard flow-matching action objective toward \(\mathbf{A}^\star\), and later serves as a stable local manipulation prior.

\paragraph{Stage 2: independent egocentric residual policy.}
In the second stage, the base policy is frozen and we add an independent residual action expert. The egocentric branch should interpret the wrist configuration in the coordinate system of the image it observes, so we express the current wrist pose in the current head frame as \(T^{H\leftarrow W}=(T^H)^{-1}T^W\), where \(T^W\) and \(T^H\) are both recorded in the shared collection/world frame. This gives the wrist pose in the head-camera frame and aligns the pose input with \(I^H\). Conditioned on the head/egocentric image and this head-frame wrist pose, the residual branch predicts a complete action candidate \(\mathbf{A}^r\) in the same wrist-relative action space as \(\mathbf{A}^b\), along with a scalar gate \(\alpha\). The final predicted action is formed by gated blending:
\begin{equation}
    T^{H\leftarrow W} = (T^H)^{-1}T^W,
    \qquad
    (\mathbf{A}^r,\alpha) = \pi_r(I^H,T^{H\leftarrow W},\ell),
    \qquad
    \hat{\mathbf{A}} = (1-\alpha)\mathbf{A}^b + \alpha\mathbf{A}^r.
    \label{eq:final_action}
\end{equation}
This coordinate transform is only used as residual input. It does not change the action target: during training, \(\mathbf{A}^r\) is supervised by the same demonstration action chunk \(\mathbf{A}^\star\) as the base policy. Thus the base action, egocentric action candidate, and final action share one coordinate convention and can be mixed directly. The gate controls how much the policy moves from the stable wrist-view base action toward the egocentric candidate.

\paragraph{Training objective.}
Since the base policy is fixed in stage 2, the residual action expert directly learns the full ground-truth action \(\mathbf{A}^\star\) rather than the difference \(\mathbf{A}^\star-\mathbf{A}^b\). We train this branch with the same flow-matching objective, and also supervise the composed action:
\begin{equation}
    \mathcal{L}_{\mathrm{GERP}}
    =
    \lambda_{\mathrm{res}}
    \mathcal{L}_{\mathrm{FM}}
    (\pi_r;\mathbf{A}^\star)
    +
    \lambda_{\mathrm{act}}
    \left\|
    (1-\alpha)\mathbf{A}^b+\alpha\mathbf{A}^r-\mathbf{A}^\star
    \right\|_2^2 .
    \label{eq:gerp_loss}
\end{equation}
The flow-matching loss teaches the egocentric branch to generate action from head/egocentric context, while the composed-action loss trains the gate to combine this candidate with the wrist-view base action. We use a curriculum schedule for stable optimization: starts with only the residual branch loss \((\lambda_{\mathrm{act}}=0)\), then linearly increases to equal weighting \((\lambda_{\mathrm{act}}=1)\), and finally continues training with this balanced objective.
At inference, GERP computes the wrist-view base action, the egocentric action candidate and the gating value, then executes the fused action in Equation~\eqref{eq:final_action}.

\begin{table*}[t]
\centering
\resizebox{0.9\textwidth}{!}{
\begin{tabular}{lcccccc}
\toprule
\textbf{Task} & \textbf{EgoGuide}
& 100 Demos & 200 Demos & 300 Demos & 400 Demos \\
\midrule
\multirow{2}{*}{Pick Cube}
& \xmark  & 25\% / 30.0\%  & 40\% / 42.5\% & 50\% / 55.0\% & 70\% / 70.0\%  \\
& \cmark  & \textbf{30\%} / \textbf{47.5\%} & \textbf{65\%} / \textbf{75.0\%} & \textbf{95\%} / \textbf{97.5\%} & \textbf{100\%} / \textbf{100.0\%} \\
\midrule
\multirow{2}{*}{Pepper Sorting}
& \xmark  & 0\% / 0.0\% & 10\% / 12.5\%  & 15\% / 45.0\%  & 50\% / 57.5\%  \\
& \cmark  & 0\% / \textbf{7.5\%} & \textbf{50\%} / \textbf{60.0\%} & \textbf{60\%} / \textbf{60.0\%}  & \textbf{75\%} / \textbf{77.5\%}  \\
\midrule
\multirow{2}{*}{Garlic Storage}
& \xmark  & 0\% / 0.0\% & 0\% / 0.0\% & 0\% / 18.8\% & 0\% / 16.3\%  \\
& \cmark  & 0\% / \textbf{23.8\%} & \textbf{15\%} / \textbf{38.8\%} & \textbf{35\%} / \textbf{53.8\%} & \textbf{50\%} / \textbf{61.3\%}  \\
\bottomrule
\end{tabular}
}
\caption{Performance scaling curves of different data collection strategies with or without EgoGuide. We report success rate and progress score as ``SR / TPS''.}
\label{tab:data_diversity_scaling}
\vspace{-3mm}
\end{table*}

\section{Experiments}
\label{sec:result}

\subsection{Experimental Setup}

\paragraph{Robot deployment.}
All policies are evaluated on a Flexiv Rizon 4 robot with a Grav gripper. We mount the fisheye camera and Meta Quest controller in the same configuration as EgoGuide-UMI, and use a fixed Meta Quest headset as the head/egocentric camera at a third-person viewpoint close to the demonstrator's viewing direction. Data collection and robot evaluation use disjoint rooms and scene layouts, while the fixed setup keeps camera intrinsics and relative 6D pose consistent with training. We run 20 trials per policy with object-location randomization.

\paragraph{Policy implementation.}
We use $\pi_{0.5}$ as the base model. Images are resized to $224\times224$, and the action horizon is $K=16$. We fine-tune the model for 30~K steps with batch size 128 and cosine learning rate from $2.5e-5$ to $2.5e-6$. For GERP, we first train the wrist-view base policy, then freeze it and train the egocentric branch for 30~K steps. The composed-action loss is introduced after a 15~K-step egocentric warm-up, linearly ramped over 10~K steps, and kept for the remaining steps.

\begin{figure*}[t]
\centering
\includegraphics[width=\textwidth]{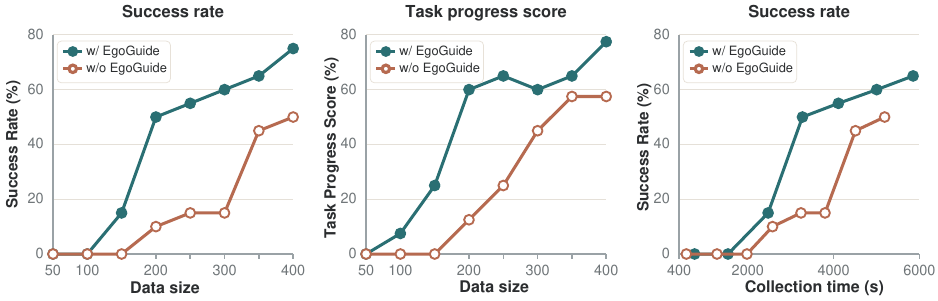}
\caption{Data scaling comparison between unguided and EgoGuide-guided collection. For each dataset size, we train the same $\pi_{0.5}$ policy and report both success rate and progress score.}
\label{fig:guidance_scaling}
\end{figure*}

\begin{figure*}[t]
    \centering
    \begin{minipage}[c]{0.27\textwidth}
        \centering
        \includegraphics[width=0.8\textwidth]{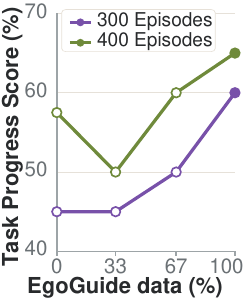}
        \caption{Mixing regular data and EgoGuide data.}
        \label{fig:mix-data}
    \end{minipage}
    \hfill
    \begin{minipage}[c]{0.7\textwidth}
        \centering
        \includegraphics[width=\textwidth]{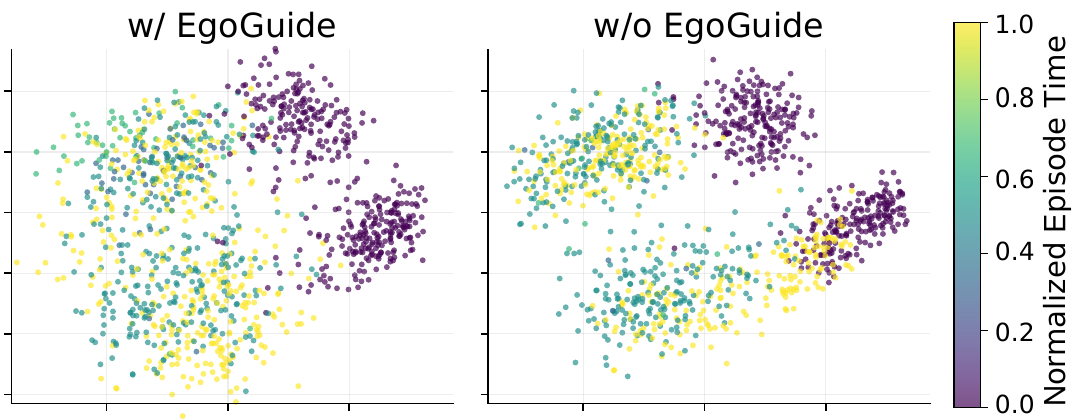}
        \caption{t-SNE visualization of wrist-camera CLIP features.}
        \label{fig:wrist-pca}
    \end{minipage}
    \vspace{-4mm}
\end{figure*}

\paragraph{Metrics and tasks.}
We report both binary success rate (SR) and task progress score (TPS). The task progress captures partial completion of subtasks. The tasks and defined subgoals are as follows:

\begin{enumerate}[nosep]
    \item \textbf{Pick Cube}: Pick up a yellow cube (TPS $=50\%$) and put it in a box (TPS $=100\%$).
    \item \textbf{Pepper Sorting}: Pick up a green pepper (TPS $=25\%$) and put it in the green tray (TPS $=50\%$), then pick up a chili pepper (TPS $=75\%$) and put it in the red tray (TPS $=100\%$).
    \item \textbf{Garlic Storage}: Open the top drawer (TPS $=25\%$), pick up the garlic (TPS $=50\%$), put it in the drawer (TPS $=75\%$), then close the drawer (TPS $=100\%$).
    \item \textbf{Rubik's Cube}: Grasp the Rubik's cube (TPS $=50\%$) and rotate $90^\circ$ (TPS $=100\%$).
\end{enumerate}

\subsection{Comparison on Online Data Curation}

We first evaluate whether EgoGuide improves the utility of collected demonstrations, using the wrist-only $\pi_{0.5}$ policy for all experiments.
For each task, users collect two datasets with the same target size: one using standard unguided collection and one using EgoGuide feedback. Each session contains 100 samples; we reset the collection location and EgoGuide memory between sessions, while keeping background, scene, and lighting matched for fair comparison.

\paragraph{Performance and efficiency comparison.} Tab.~\ref{tab:data_diversity_scaling} and Fig.~\ref{fig:guidance_scaling} show that EgoGuide improves policy performance and reduces the required data. On Pepper Sorting with 200 demonstrations, for example, EgoGuide raises success from 10\% to 50\% and reaches 50\% success using only half the data. The gains also appear on long-horizon tasks, even though EgoGuide only guides collection starts.
EgoGuide largely preserves UMI collection efficiency: real-time guidance adds 4.3~s per sample on average, but the total-time scaling curve in Fig.~\ref{fig:guidance_scaling} remains better than unguided collection.
Fig.~\ref{fig:mix-data} further shows that mixing more EgoGuide data into a regular dataset steadily improves performance, indicating consistent data quality.

\paragraph{Dataset distribution.} We visualize camera image features to compare diversity and coverage. Fig.~\ref{fig:wrist-pca} shows CLIP wrist-image features as an example: EgoGuide covers a larger feature-space region. Measured by feature covariance trace, EgoGuide improves variance by 5\%, 4\%, 3\%, and 4\% across the two camera views and two feature models.

\begin{figure*}[t]
    \centering
    \begin{minipage}[c]{\textwidth}
        \centering
        \includegraphics[width=0.65\textwidth]{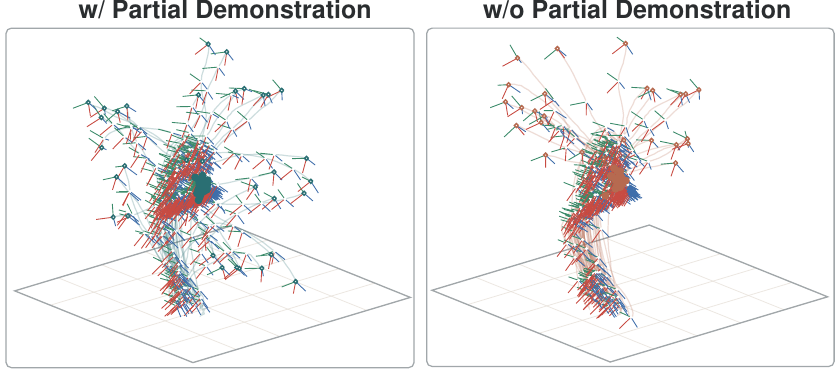}
        \vspace{-1mm}
        \caption{Trajectory distribution of EgoGuide with and without partial demonstration (25 episodes each). Partial demonstration substantially increases data coverage.}
        \label{fig:partial-demo}
    \end{minipage}
    \vspace{-3mm}
\end{figure*}

\paragraph{Partial demonstration.} Fig.~\ref{fig:partial-demo} compares EgoGuide with and without the partial-demonstration option described in Sec.~\ref{sec:data-eval}. Mid-task starts prevent trajectories from collapsing to similar later states and substantially increase spatial coverage.

\subsection{Egocentric Residual Policy Evaluation}

We compare GERP with \emph{Wrist Only}, the default UMI setting, and \emph{Wrist+Ego Direct}, which treats the egocentric camera as a regular head-view input and feeds both camera views to the policy.
Tab.~\ref{tab:gerp_main} shows that on Pepper Sorting, GERP improves success rate and progress score by 5\%--10\% over \emph{Wrist Only}. \emph{Wrist+Ego Direct} can sometimes degrade performance because the moving egocentric view mismatches policy pretraining and distracts training, while GERP uses it as a gated residual cue.
We also visualize the gating value throughout the rollout process in the appendix.

\begin{table*}[t]
\centering
\resizebox{0.85\textwidth}{!}{
\begin{tabular}{lc|ccc}
\toprule
\textbf{Task} & \textbf{\#Demos} & \textbf{Wrist Only} & \textbf{Wrist+Ego Direct} & \textbf{GERP} \\
\midrule
Pick Cube           &  200  & 65\% / \underline{75.0\%} & \underline{70\%} / \underline{75.0\%} & \textbf{80\%} / \textbf{90.0\%}  \\
Pepper Sorting      &  400  & \underline{75\%} / \underline{77.5\%} & 65\% / 72.5\% & \textbf{80\%} / \textbf{87.5\%}  \\
Garlic Storage      &  400  & \underline{50\%} / 61.3\% & \underline{50\%} / \textbf{72.5\%}  & \textbf{55\%} / \underline{70.0\%} \\
Rubik's Cube        &  300  & 30\% / 37.5\% & \underline{70\%} / \underline{77.5\%} & \textbf{80\%} / \textbf{82.5\%}  \\
\bottomrule
\end{tabular}
}
\caption{Comparison of egocentric policies. We report ``SR / TPS''. Wrist+Ego Direct feeds wrist and head/egocentric images directly into $\pi_{0.5}$. Best results are bold and second-best are underlined.}
\label{tab:gerp_main}
\vspace{-5mm}
\end{table*}


\section{Conclusion}

We present EgoGuide, a robot-free demonstration collection system that augments UMI-style data with synchronized wrist and head/egocentric observations. Its online coverage guidance and partial-demonstration support help demonstrators collect more diverse and useful trajectories, while static filtering removes common sensor failures. We also introduce GERP, which uses the head/egocentric view as a gated residual action cue on top of a stable wrist-view base policy. Together, EgoGuide and GERP improve data efficiency and policy robustness in real-world manipulation.


\clearpage


\bibliography{main}  

\clearpage

\appendix

\section{Implementation Details}

\subsection{Task Details}
\label{app:task_details}

We provide additional details on the evaluation randomization protocol. For each trial, objects are manually reset within predefined ranges while ensuring reachability and avoiding initial object overlap. All methods on the same task are evaluated under the same randomization protocol. Once execution starts, no human correction is allowed unless a safety stop is required.

\begin{itemize}
\item \textbf{Pick Cube.}
The cube position and the target box are randomized within a tabletop region of approximately 30 cm $\times$ 30 cm. Its yaw orientation is sampled within 45 degrees.

\item \textbf{Pepper Sorting.}
The green pepper and chili pepper and the two trays are independently randomized within a region of approximately 40 cm $\times$ 40 cm. Their yaw orientations are randomized.

\item \textbf{Garlic Storage.}
The garlic position is randomized within approximately 20 cm $\times$ 40 cm. The drawer cabinet is always facing the robot and shifting left to right within 40cm. The drawer always starts from a closed state.

\item \textbf{Rubik's Cube Rotation.}
The cube position is randomized within approximately 30 cm $\times$ 30 cm. Its yaw orientation is sampled within 20 degrees. The blue side always faces upward.

\end{itemize}

A trial terminates when the task succeeds, the end effector deviates from regular working area for 5 seconds, the object becomes unrecoverable, or the robot triggers a safety stop. Safety-stop trials are counted as failures.

\subsection{Data Collection Statistics}
\label{app:data_statistics}

Tab.~\ref{tab:data_stats} summarizes the datasets after synchronization, static filtering, and downsampling. We report both episode-level and frame-level statistics.

\begin{table}[ht]
\centering
\resizebox{\linewidth}{!}{
\begin{tabular}{lc|cccccc}
\toprule
Task & EgoGuide & \# Episodes & \# Frames & Frames per Episode & Duration per Episode (s) & Rejected Episodes & Scenes \\
\midrule
Pick Cube      & \xmark & 400 & 25,434 & 63.59  &  3.18 & --  & 4 \\
Pick Cube      & \cmark & 400 & 35,473 & 88.68  &  4.43 & 22 & 4 \\
\midrule
Pepper Sorting & \xmark & 400 & 50,410 & 126.03 &  6.30 & --  & 4 \\
Pepper Sorting & \cmark & 394 & 45,189 & 114.69 &  5.73 & 24 & 4 \\
\midrule
Garlic Storage & \xmark & 400 & 82,597 & 206.49 & 10.32 & --  & 4 \\
Garlic Storage & \cmark & 400 & 60,712 & 151.78 &  7.59 & 9  & 4 \\
\bottomrule
\end{tabular}
}
\caption{Dataset statistics after synchronization, filtering, and downsampling. ``Frames'' denotes synchronized observation-action frames used for training.}
\label{tab:data_stats}
\end{table}

\begin{figure}[t]
\centering
\begin{subfigure}[b]{0.48\textwidth}
    \centering
    \includegraphics[width=\textwidth]{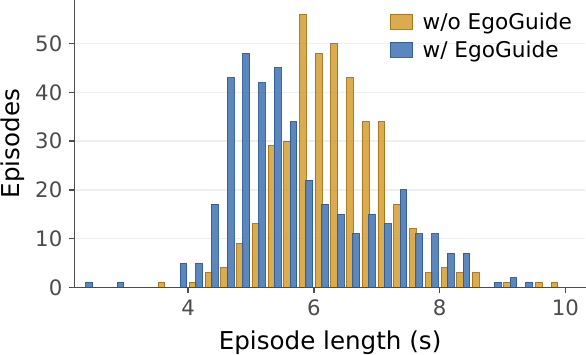}
    \caption{Pepper Sorting}
\end{subfigure}
\hfill
\begin{subfigure}[b]{0.48\textwidth}
    \centering
    \includegraphics[width=\textwidth]{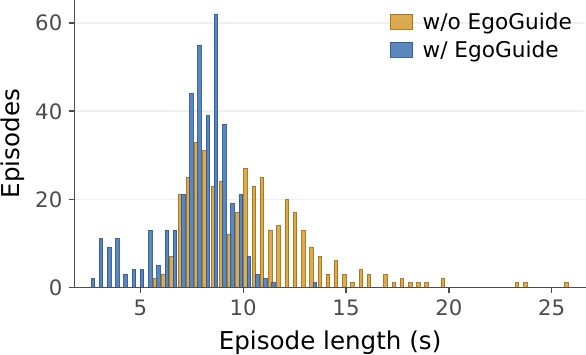}
    \caption{Garlic Storage}
\end{subfigure}
\caption{Episode-length distribution after synchronization. EgoGuide includes both full and partial demonstrations, resulting in a broader frame-count distribution.}
\label{fig:episode_length_distribution}
\end{figure}

Fig.~\ref{fig:episode_length_distribution} compares the episode-length distributions measured by the number of synchronized frames per episode. EgoGuide produces a broader distribution because the dataset contains both full demonstrations and partial demonstrations starting from intermediate states. This increases coverage of later-stage interactions and recovery behaviors that are less frequent in standard full-trajectory collection.

\subsection{Training Details}
\label{app:training_details}

All datasets are split at the episode level, with a validation ratio of 3\%. This avoids temporally adjacent frames from the same trajectory appearing in both splits. Each training sample contains an observation history of 1 frame for camera inputs and 2 frames for low-dim inputs, and an action chunk of length $K=16$.

The action target follows the wrist-relative convention and consists of relative translation, relative rotation, and gripper command. Translation, rotation, and gripper targets are normalized using statistics from the training set. Episodes are not downsampled and keep 20 Hz at training.

For all experiments, we train the $\pi_{0.5}$-based policy on 4 NVIDIA H100 GPUs and 64-core CPU. A 30~K-step training usually takes
5 hours for wrist-only policies, and 3 hours for additional GERP residual policy. We use the checkpoints of last iteration for evaluation. The evaluation is on a workstation with one NVIDIA 4070 GPU and Intel i9 7900X CPU.

\begin{figure}[t]
\centering
\includegraphics[width=0.95\linewidth]{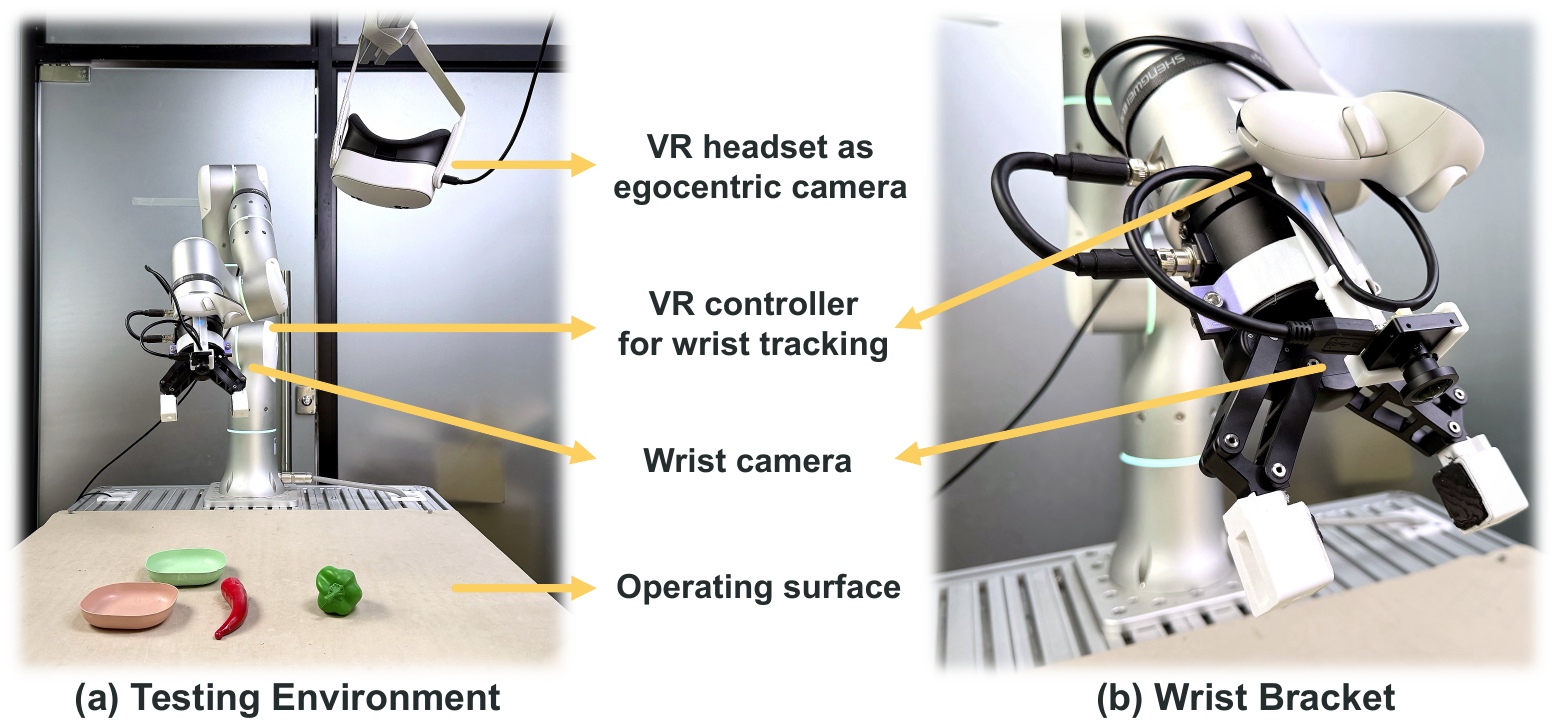}
\caption{Robot deployment setup. The wrist camera is fixed to the robot end-effector with a custom bracket. For egocentric-policy experiments, an additional fixed egocentric camera provides global scene context.}
\label{fig:robot_setup}
\end{figure}

Fig.~\ref{fig:robot_setup} shows the robot deployment setup and the wrist-camera mounting bracket. The wrist camera is rigidly attached to the robot end-effector using a fixed custom bracket. The bracket keeps the camera pose stable relative to the gripper and approximately matches the wrist-camera viewpoint used during robot-free UMI data collection.

For egocentric-policy experiments, the egocentric camera is fixed in the workspace rather than actively controlled. It is placed at approximately 70 cm height from the workspace desktop. This provides a stable global view while avoiding major occlusion from the robot arm.

The policy runs at 10 Hz. Predicted actions are clipped by predefined translation, rotation, gripper, and workspace limits before execution, and then converted to joint control signals by the internal inverse-kinematics tool of FlexivRDK. Each trial starts from the same robot home pose, followed by task-specific scene randomization.

\section{Additional Comparisons}

\begin{figure}[t]
    \centering
    \begin{subfigure}[b]{0.48\textwidth}
        \centering
        \includegraphics[width=\textwidth]{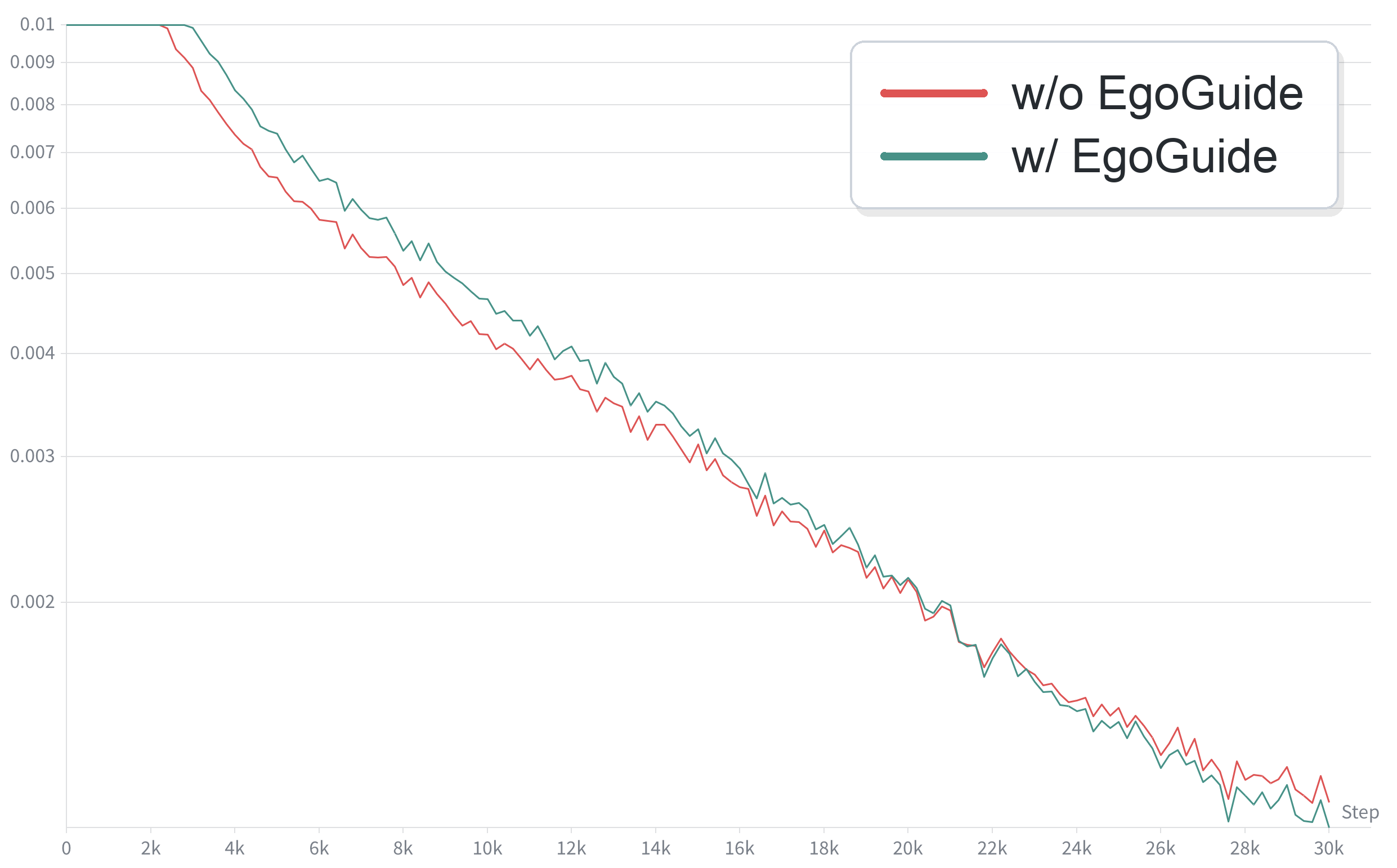}
        \caption{Pepper Sorting, 400 samples}
    \end{subfigure}
    \hfill
    \begin{subfigure}[b]{0.48\textwidth}
        \centering
        \includegraphics[width=\textwidth]{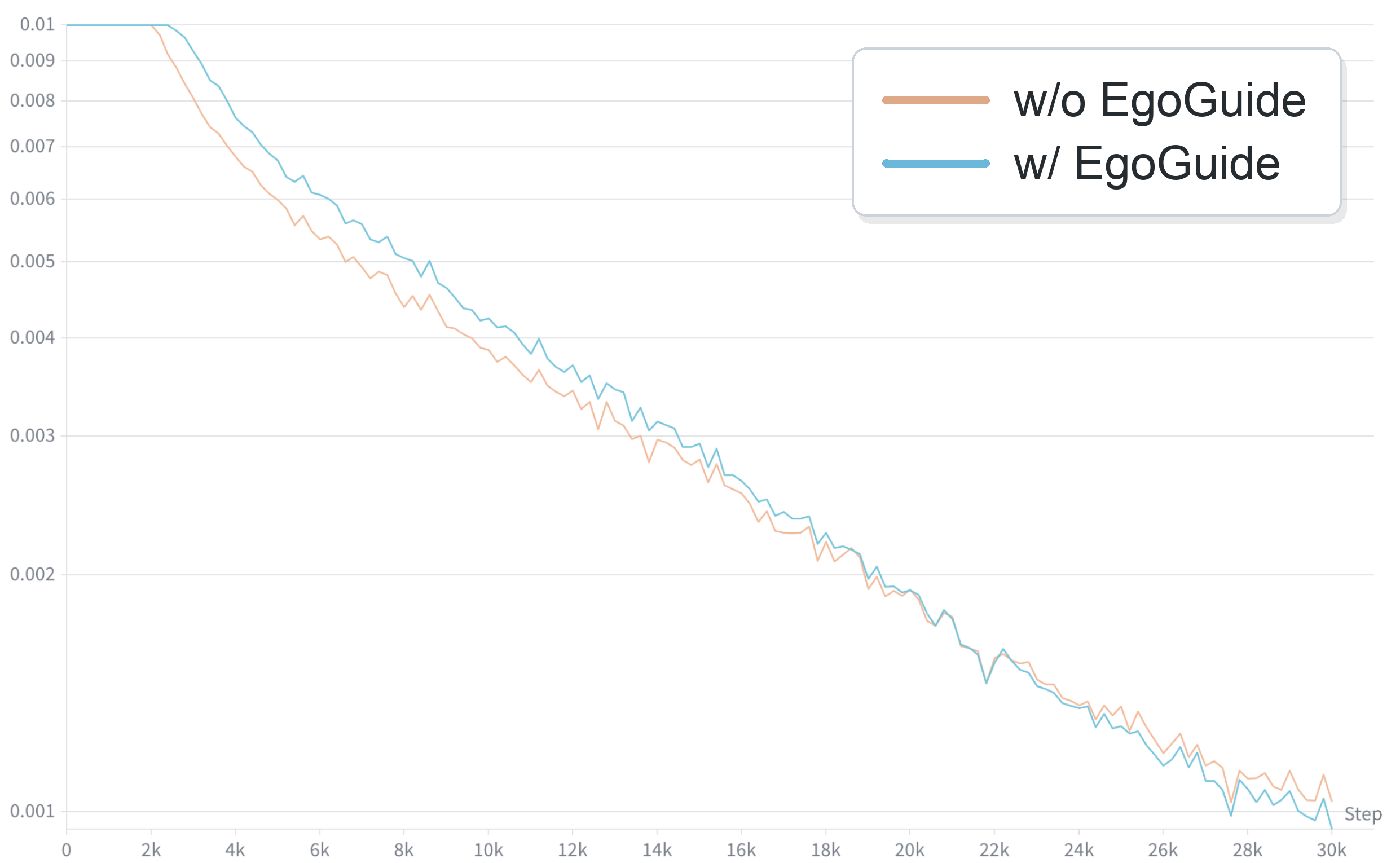}
        \caption{Pepper Sorting, 300 samples}
    \end{subfigure}
    
    \vspace{0.4cm}
    
    \begin{subfigure}[b]{0.48\textwidth}
        \centering
        \includegraphics[width=\textwidth]{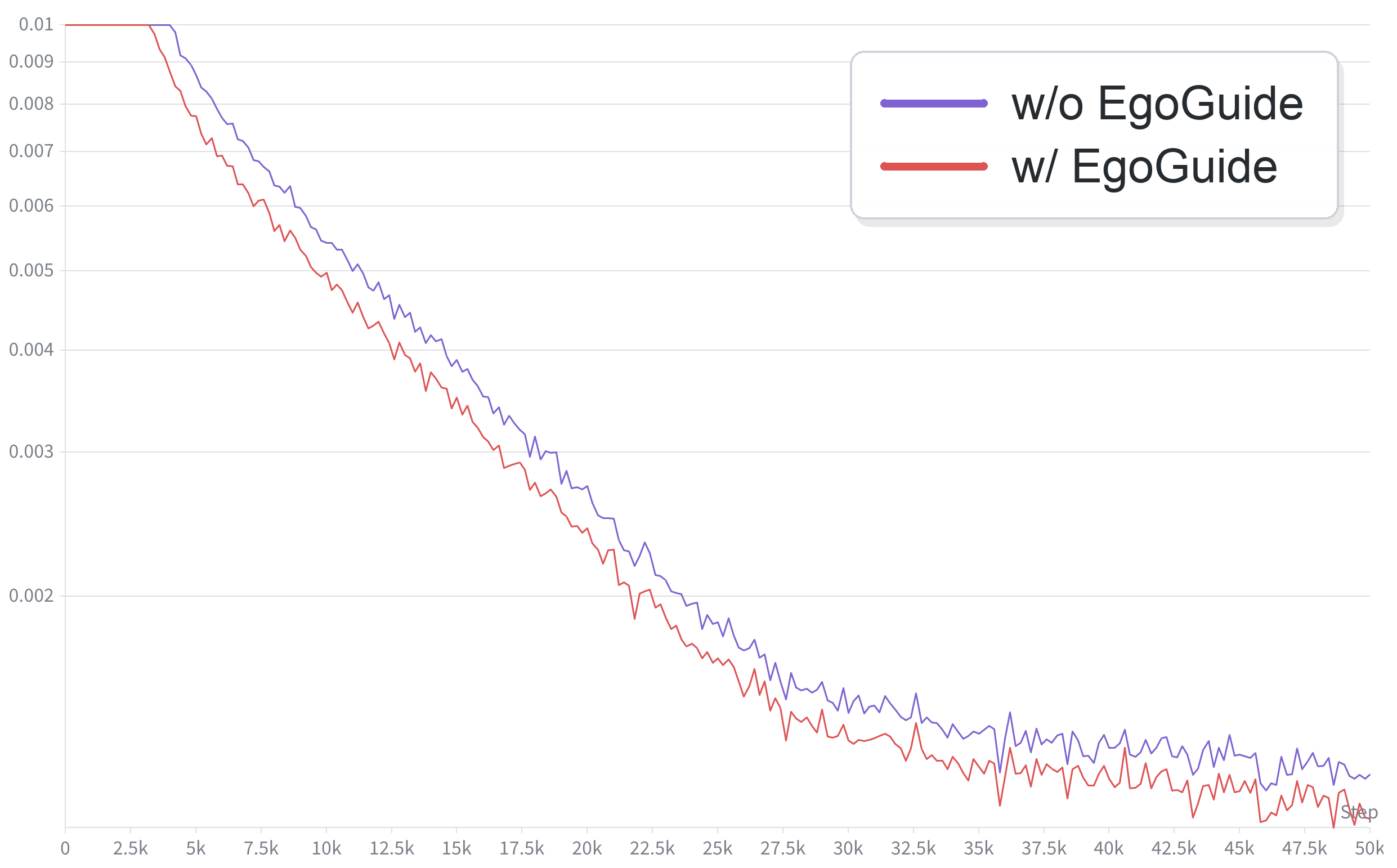}
        \caption{Garlic Storage, 400 samples}
    \end{subfigure}
    \hfill
    \begin{subfigure}[b]{0.48\textwidth}
        \centering
        \includegraphics[width=\textwidth]{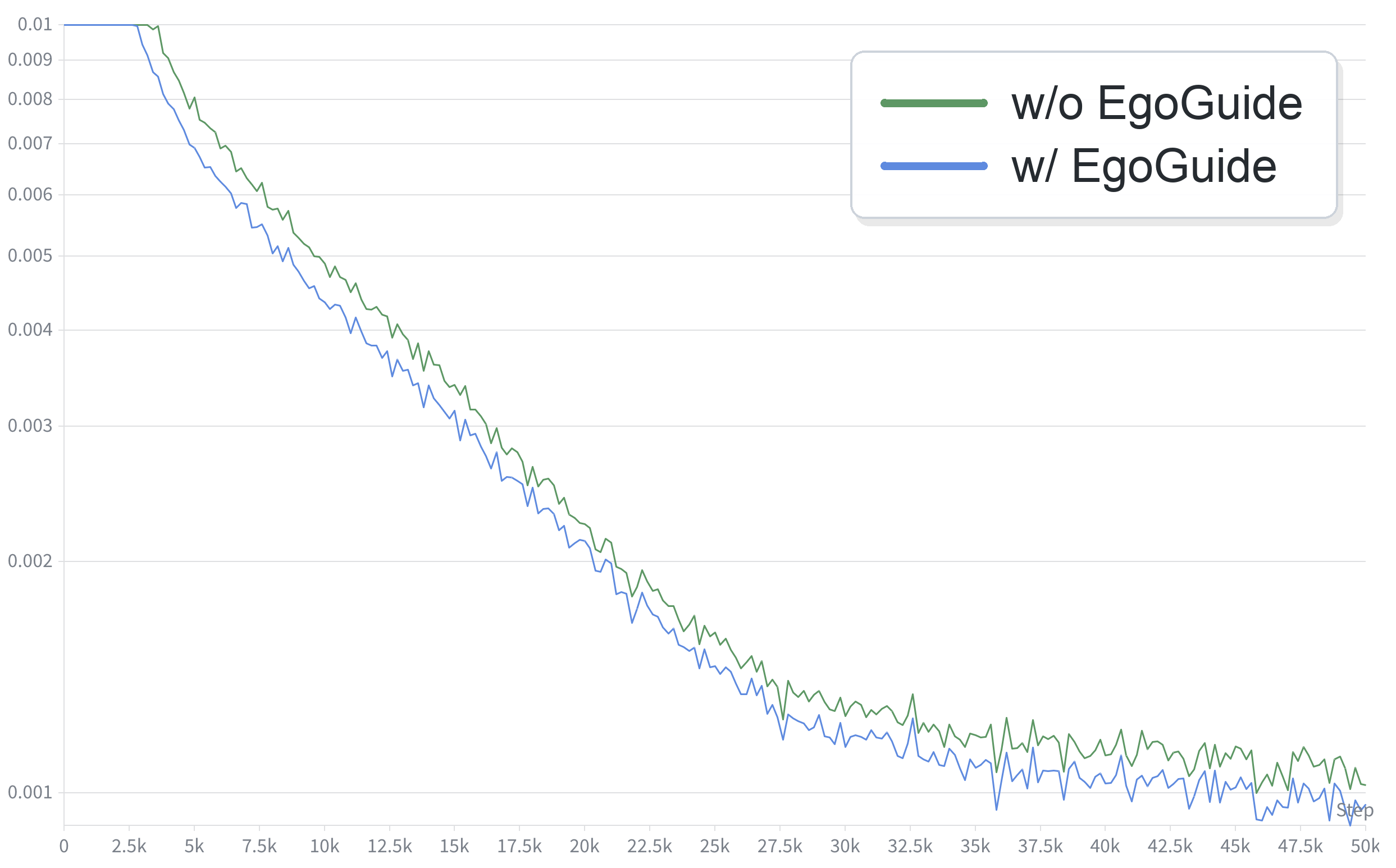}
        \caption{Garlic Storage, 300 samples}
    \end{subfigure}
    \caption{Examples of loss curves for various tasks and data scale. The loss values (Y-axis) are truncated and presented in log-scale for better visualization.}
    \label{fig:loss-cuves}
\end{figure}

\begin{table}[t]
    \centering
    \begin{tabular}{lccc}
        \toprule
        Method & Setting 1 & Setting 2 & Setting 3 \\
        \midrule
        Wrist+Ego Direct & 65\% / 72.5\% & 60\% / 65.0\% & 40\%/ 40.0\% \\
        GERP              & 80\% / 87.5\% & 75\% / 80.0\% & 70\% / 80.0\% \\
        \bottomrule
    \end{tabular}
    \caption{Pepper Sorting performance under different fixed egocentric camera placements. Each entry reports success rate and task progress score as SR / TPS.}
    \label{tab:head_camera_positions}
\end{table}

\subsection{Loss Behaviors}

Although the training loss does not directly reflect the final model performance or task success rate, it may still provide useful indications about the quality of the post-training data. 
As shown in Fig.~\ref{fig:loss-cuves}, we present the post-training loss curves on the \textit{Pepper Sorting} and \textit{Garlic Storage} tasks. 
In the early stage of post-training, the loss decreases differently when trained with data of varying quality. 
However, in the later stage, the model trained with EgoGuide data generally converges to a lower loss, which is consistent with EgoGuide providing more informative training data.
For \textit{Pepper Sorting} task, the model trained without EgoGuide initially shows faster convergence. 
This may be because the non-EgoGuide data is more concentrated and easier to fit at the beginning; nevertheless, its loss soon saturates and fails to decrease further, while EgoGuide enables continued optimization and achieves a lower final convergence loss.

\subsection{Sensitivity to Fixed Egocentric Camera Placement}
\label{app:fixed_head_camera}

We further evaluate the robustness of egocentric policies to different fixed head-camera placements on the Pepper Sorting task. During testing, the wrist-camera and robot setup and policy checkpoint are kept unchanged; only the fixed egocentric camera pose is varied. We compare under three camera placements in Tab.~\ref{tab:head_camera_positions}. Across different camera placements, GERP is more stable than the direct Wrist-Ego baseline.

\section{Visualizations}

\subsection{Visualization of Gating Behavior}

\begin{figure*}[t]
\centering
\includegraphics[width=0.99\linewidth]{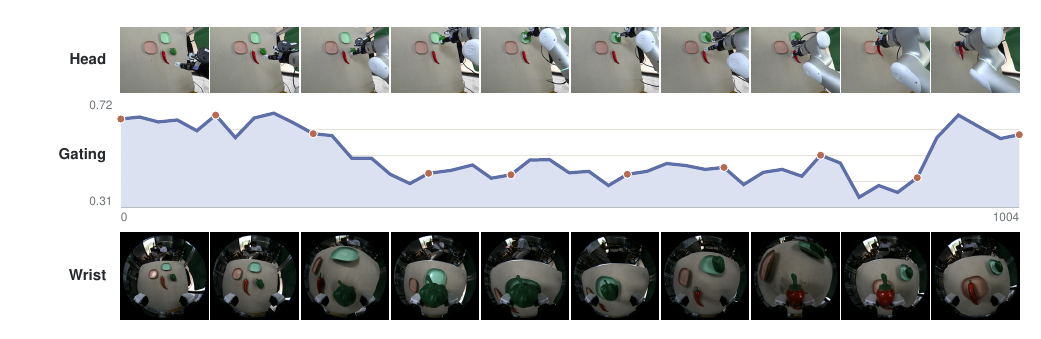}
\caption{Gating behavior during execution.}
\label{fig:gate_visualization}
\end{figure*}

Fig.~\ref{fig:gate_visualization} shows the gating value during evaluation. Notably, the gating value increases when the target object is occluded or missing from the wrist view (right side of the figure).

\subsection{Additional Visualization of Feature Distribution}

\begin{figure}[h]
    \centering
    \begin{subfigure}[b]{0.9\textwidth}
        \centering
        \includegraphics[width=\textwidth]{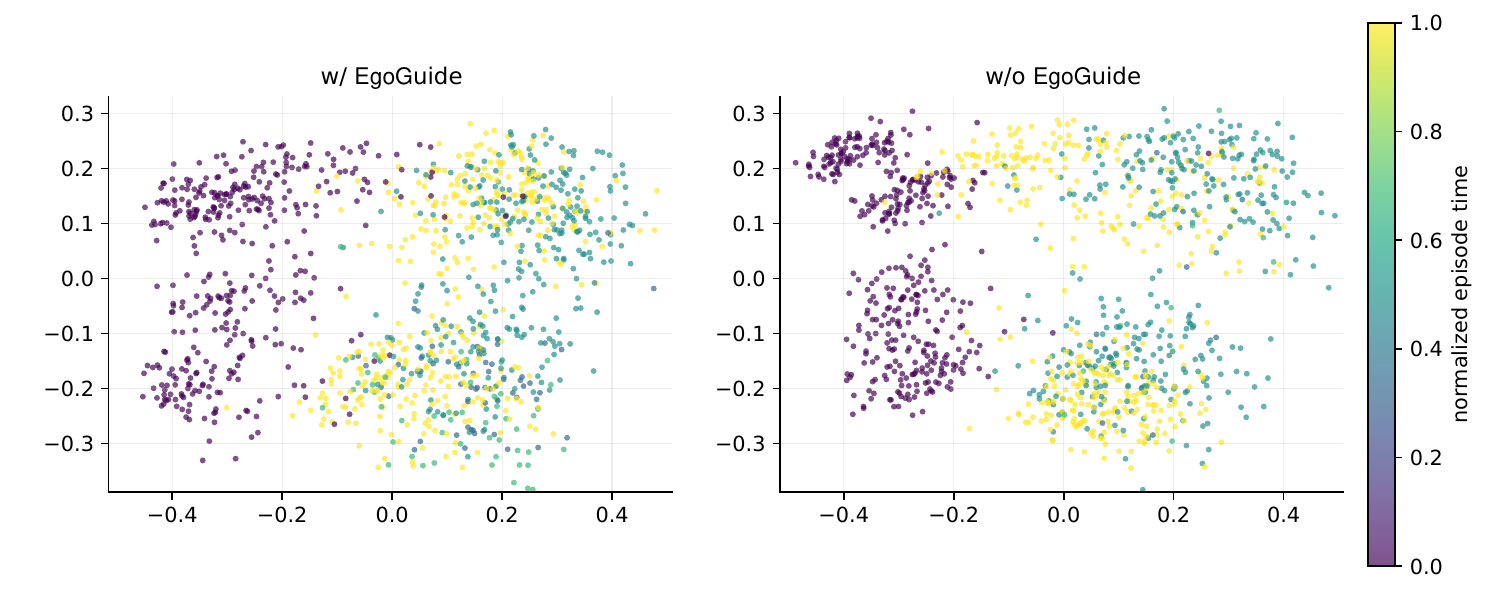}
        \caption{Wrist camera, DINO feature.}
    \end{subfigure}
    \begin{subfigure}[b]{0.9\textwidth}
        \centering
        \includegraphics[width=\textwidth]{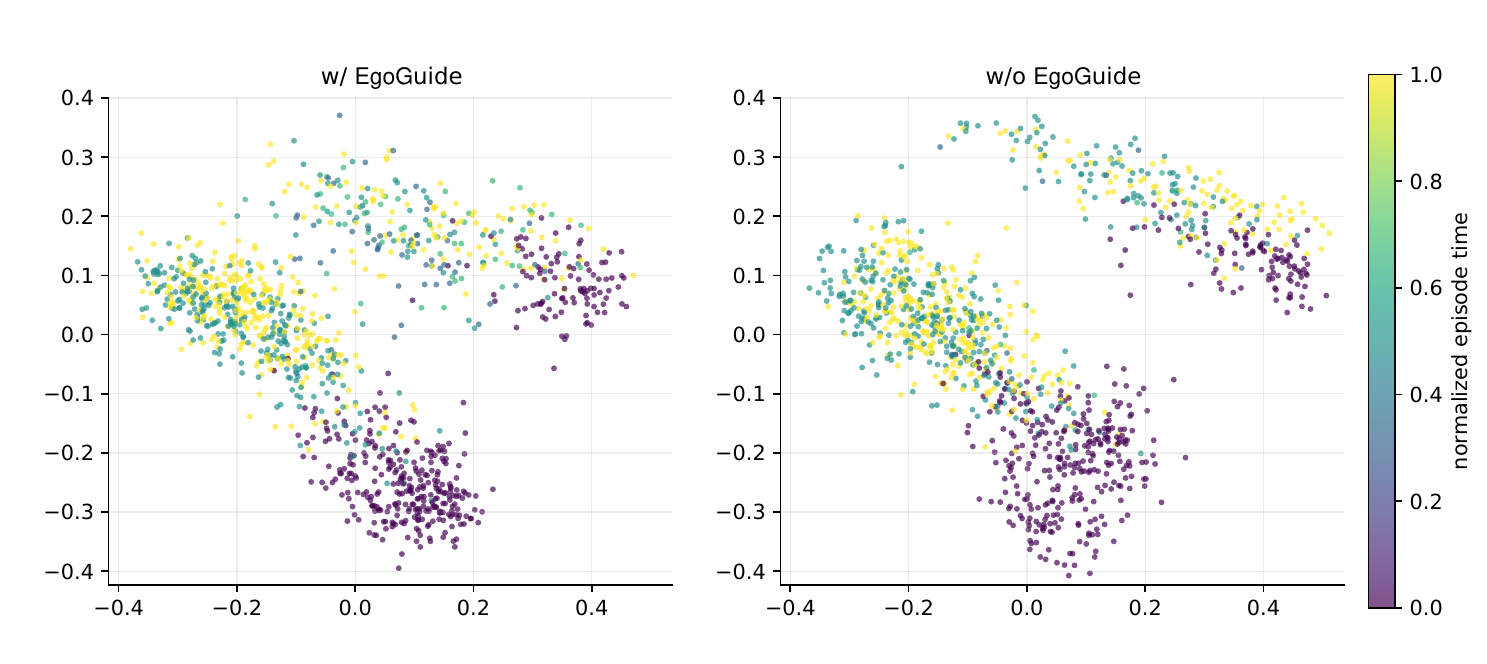}
        \caption{Egocentric camera, CLIP feature.}
    \end{subfigure}
    \begin{subfigure}[b]{0.9\textwidth}
        \centering
        \includegraphics[width=\textwidth]{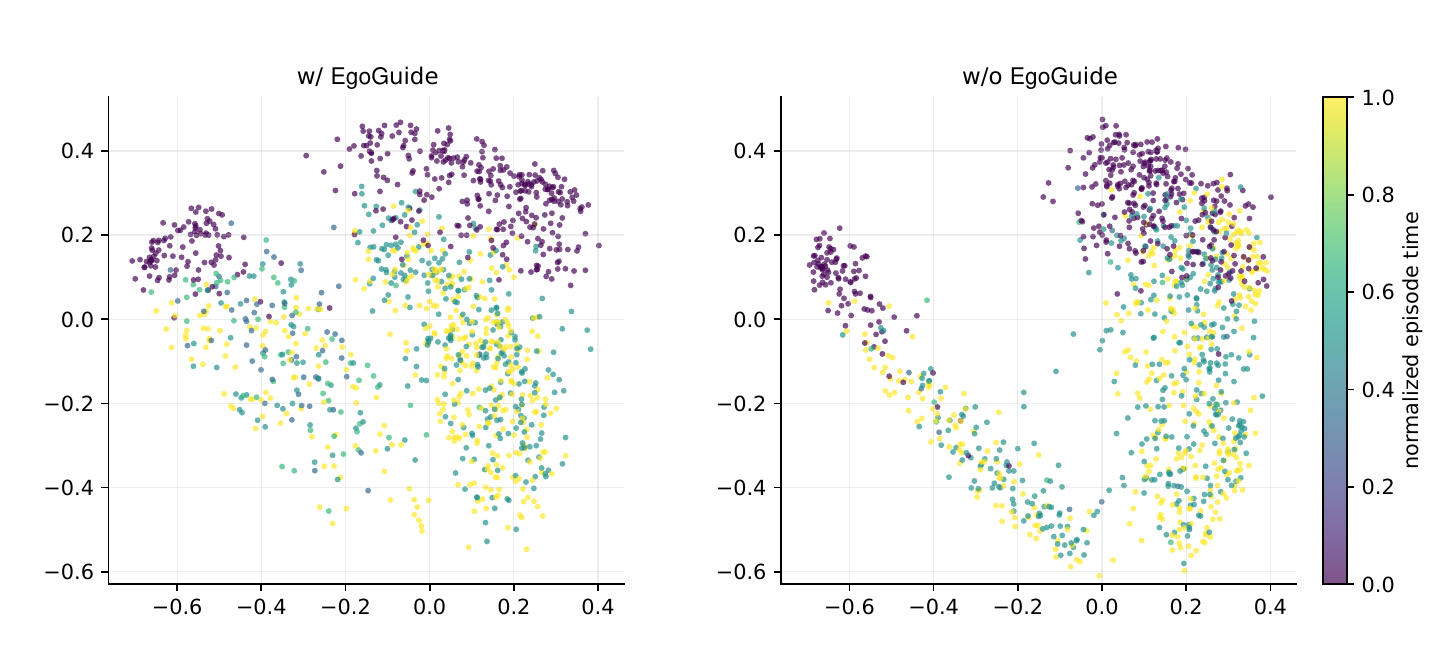}
        \caption{Egocentric camera, DINO feature.}
    \end{subfigure}
    \caption{t-SNE visualization of feature distributions. EgoGuide enhances data diversity and unguided data is more concentrated in fewer regions of the projected feature space.}
    
    \label{fig:tsne_appendix}
\end{figure}

Fig.~\ref{fig:tsne_appendix} (on the next page) presents additional t-SNE visualizations for all other combinations of camera modalities (wrist and egocentric) and feature encoders (CLIP and DINOv2) on Pepper Sorting task datasets. Similar to the main-paper result, EgoGuide data consistently exhibits broader feature-space coverage across all settings. The effect is observed in both local wrist-view observations and global egocentric observations, indicating that the online guidance mechanism increases diversity in both manipulation-centric and scene-level visual states.

\subsection{Qualitative Data Comparison}

\begin{figure}[ht]
    \centering
    \begin{subfigure}[b]{0.48\textwidth}
        \centering
        \includegraphics[width=\textwidth]{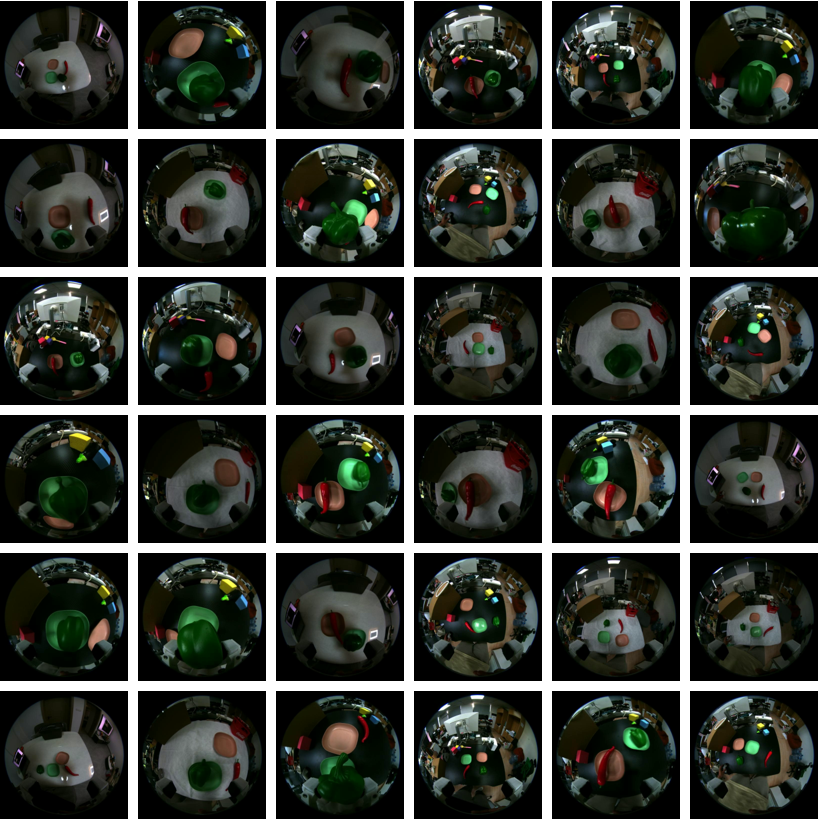}
        \caption{Wrist camera, w/o EgoGuide}
    \end{subfigure}
    \hfill
    \begin{subfigure}[b]{0.48\textwidth}
        \centering
        \includegraphics[width=\textwidth]{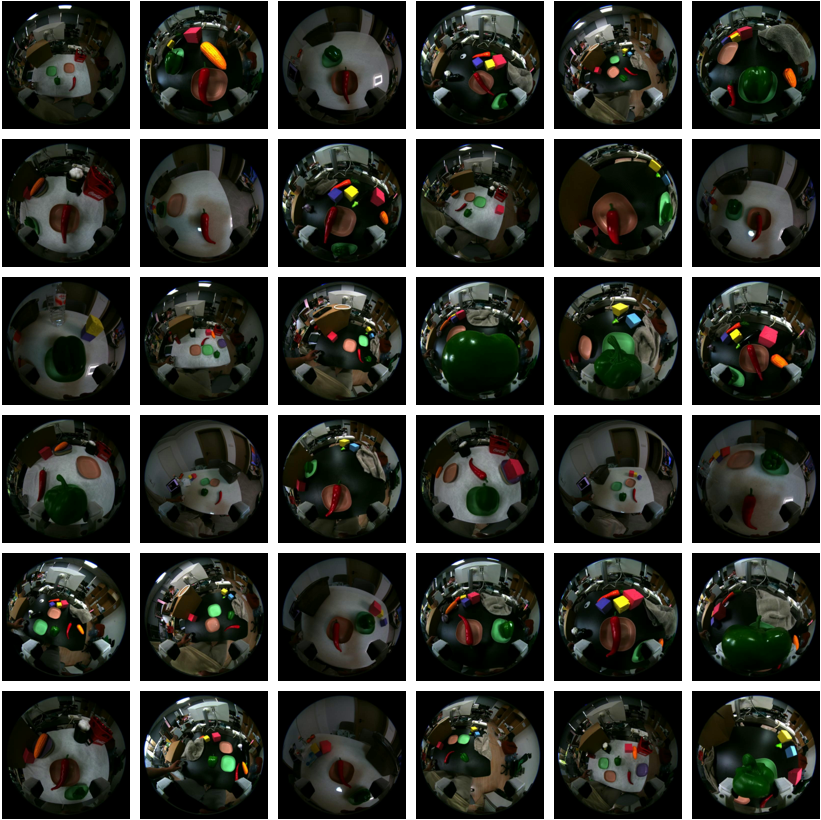}
        \caption{Wrist camera, w/ EgoGuide}
    \end{subfigure}
    
    \vspace{0.4cm}
    
    \begin{subfigure}[b]{0.48\textwidth}
        \centering
        \includegraphics[width=\textwidth]{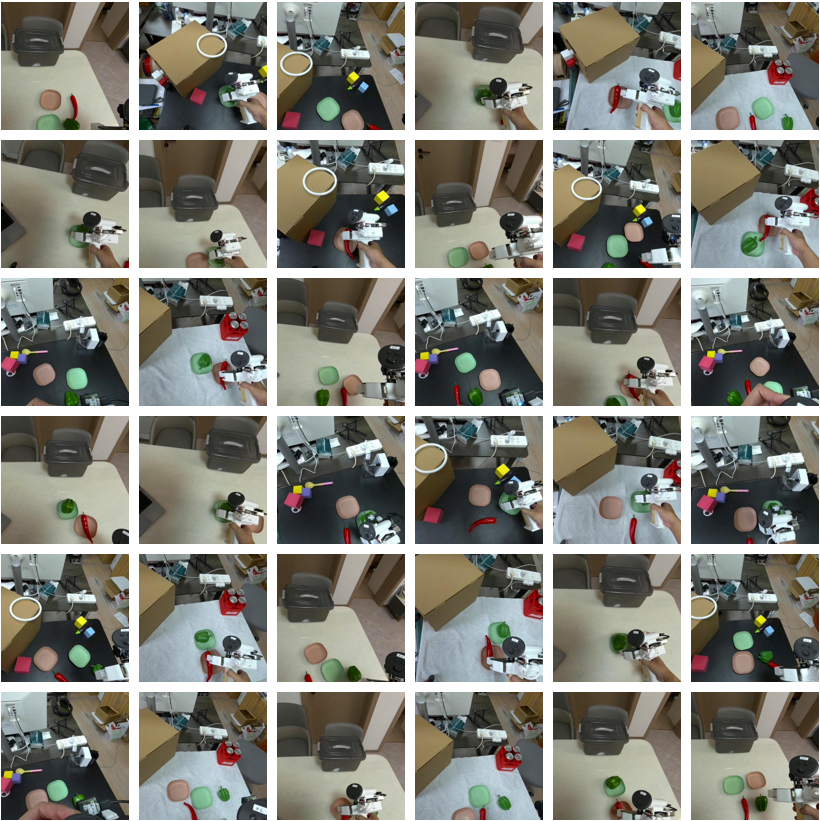}
        \caption{Egocentric camera, w/o EgoGuide}
    \end{subfigure}
    \hfill
    \begin{subfigure}[b]{0.48\textwidth}
        \centering
        \includegraphics[width=\textwidth]{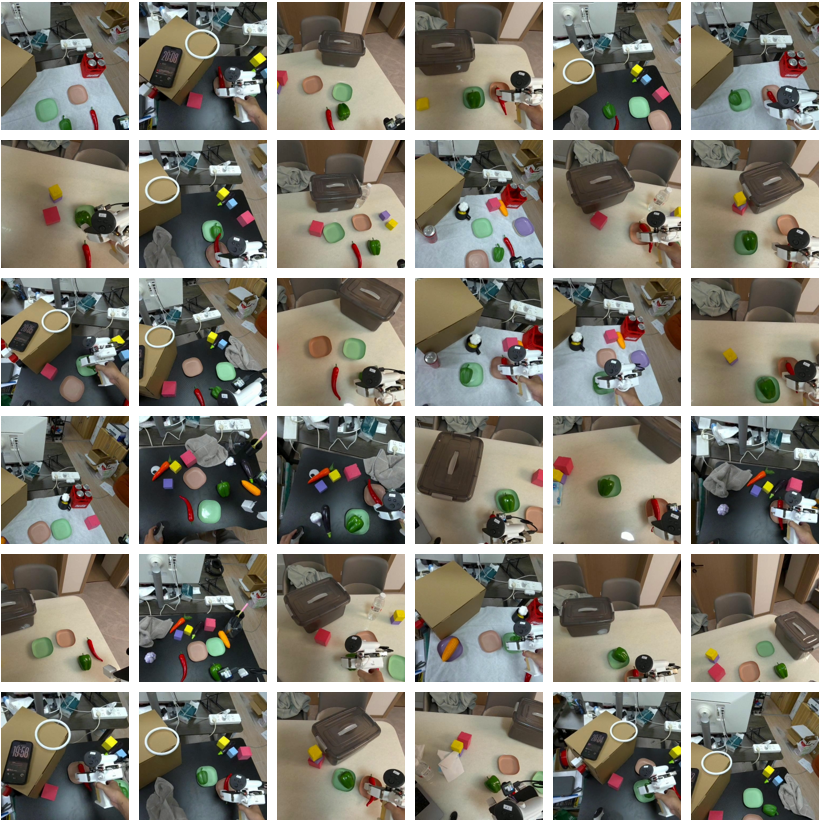}
        \caption{Egocentric camera, w/ EgoGuide}
    \end{subfigure}
    \caption{Examples of collected episodes for Pepper Sorting task.}
    \label{fig:example-image}
\end{figure}

Fig.~\ref{fig:example-image} (on the next two pages) compares representative demonstrations collected with and without EgoGuide. Samples collected with EgoGuide exhibit greater variation in viewpoints, object placements, hand poses, and scene configurations, whereas unguided collection tends to produce more repetitive observations. These qualitative results are consistent with the feature-space analysis and support our claim that EgoGuide improves dataset diversity by encouraging exploration of underrepresented states during collection.

\section{Discussion}

\textbf{Q1: Why can guiding the initial state improve the diversity of the whole episode?}
Although EgoGuide provides guidance before recording, the initial state is a strong causal factor for the rest of the trajectory. Different initial hand poses, object layouts, and viewpoints lead to different approach directions, contact sequences, occlusion patterns, and recovery behaviors. Therefore, improving the coverage of initial visual-geometric states is a simple but effective way to induce more diverse full demonstrations, without requiring policy rollouts or robot interaction. Furthermore, the introduction of \textbf{partial demonstration} indicates that any intermediate state of a task could be regarded as an \textit{initial state} for data collection, which allows controlling the data quality of the whole episode.

\textbf{Q2: Does partial demonstration introduce more late-stage samples, thus result in data imbalance?}
Yes. Partial demonstration intentionally increases the coverage of intermediate and late-stage states. However, more data samples do not necessarily indicate training imbalance. In long-horizon tasks, full demonstrations often share similar later-stage configurations, so simply collecting more complete episodes may still leave these regions under-diversified. Therefore the late stages are harder to learn so more samples for late stages could help to train a temporally balanced model. 

\textbf{Q3: Why not use DAgger for data quality?}
DAgger is designed for a different setting: it requires a learned policy to roll out, visit its own states, and query expert corrections. EgoGuide targets the earlier robot-free data collection stage, where a reliable policy may not yet exist and where on-robot interaction is undesirable. Our goal is therefore training-free data improvement: provide online collection guidance before policy training, rather than relying on policy-dependent data aggregation.

\textbf{Q4: Why use spatial wrist pose for EgoGuide instead of object-relative pose?}
The coverage score is used to detect whether the current collection state is redundant, not to define the final control representation. Spatial wrist pose is useful because it reflects action-side geometric diversity: repeated wrist poses often correspond to repeated approach directions and similar supervised end-effector motions. Object-relative pose could be informative, but it would require reliable object tracking or pose estimation in unconstrained robot-free scenes. EgoGuide instead combines wrist pose with wrist-view and egocentric visual features, which keeps the interface simple while still capturing both geometry and scene context.

\textbf{Q5: Why is EgoGuide especially suitable for crowdsourced collection?}
Crowdsourcing can scale robot-free demonstration collection, but it also makes quality control harder because users differ in skill and recording behavior. And the crowdsourced demonstrators may unintentionally produce repetitive, low-diversity, or shortcut-like demonstrations. EgoGuide addresses this with lightweight online feedback and deterministic post-recording checks. The AR novelty score tells users what kind of state is worth collecting, while static filtering removes common failures such as missing modalities, blur, abnormal brightness, and implausible pose jumps. This makes the collection protocol more standardized without requiring expert supervision for every episode.

\section{Limitation}

\paragraph{Overhead of EgoGuide.} Data collection with EgoGuide requires additional time for feature-memory computation (about 2~s) and scene and UMI pose adjustment (about 3~s). The feature computation could be further optimized by engineering efforts.

\paragraph{Ergonomics concerns.} Our volunteers report fatigue after collecting data with an AR headset. We currently alleviate this by dividing dataset collection into short sessions with sufficient rest, and we expect the headset to be replaced by AR glasses in the future.

\end{document}